\documentclass[runningheads]{llncs}
\usepackage{graphicx}
\usepackage{tabularx}
\usepackage{makecell}
\usepackage{hyperref}
\begin{document}

\title{DEEPAGÉ: Answering Questions in Portuguese about the Brazilian Environment}
\titlerunning{DEEPAGÉ: Answering Questions in Portuguese}
\author{Flávio Nakasato Cação\inst{1,\dagger}\orcidID{0000-0003-4771-6009} \and
        Marcos Menon José\inst{1,\dagger}\orcidID{0000-0003-4663-4386} \and
        André Seidel Oliveira\inst{1}\orcidID{0000-0001-6551-6911} \and
        Stefano Spindola\inst{1}\orcidID{0000-0002-3604-9346} \and
        Anna Helena Reali Costa\inst{1}\orcidID{0000-0001-7309-4528} \and
        Fábio Gagliardi Cozman\inst{1}\orcidID{0000-0003-4077-4935}}
        
\authorrunning{F. Cação et al.}

\institute{Escola Politécnica, Universidade de São Paulo, Sao Paulo, Brazil \\
$\dagger$ These authors contributed equally to the work \\
\email{\{flavio.cacao, marcos.jose, andre.seidel, stefano.spindola, anna.reali, fgcozman\}@usp.br}}

\maketitle              % typeset the header of the contribution

\begin{abstract}
The challenge of climate change and biome conservation is one of the most 
pressing issues of our time --- particularly in Brazil, where key environmental reserves 
are located. Given the availability of large textual databases on ecological themes, it is natural to resort to question answering (QA) systems to increase social awareness and understanding about these topics. In this work, we introduce multiple QA systems that combine in novel ways the BM25 algorithm, a sparse retrieval technique, with PTT5, a pre-trained state-of-the-art language model. Our QA systems focus on the Portuguese language, thus offering resources not found elsewhere in the literature. As training data, we collected questions from open-domain datasets, as well as content from the Portuguese Wikipedia and news from the press. We thus contribute with innovative architectures and novel applications, attaining an F1-score of 36.2 with our best model.

\end{abstract}

\section{Introduction}
Developing technologies and public policies to address the challenges of climate change is a multifaceted task. The United Nations (UN) Resolution with 17 Sustainable Development Goals, whose focus is on the protection of the global environment and increased general prosperity, contains a call for ``Climate Action", with five urgent goals to fight climate change; raising collective awareness belongs to these goals. Brazil has a central role in that debate as it has some of the richest and most important biomes in the world, and at the same time, it faces difficulties in curbing deforestation and illegal forest fires~\cite{West2019}. There is much to be done still in informing a poorly educated but interested population on issues related to the environment \cite{Severo2021}.

We can take advantage of the latest advances in natural language processing (NLP) and question answering (QA) so as to better inform the population on environmental issues. 
While natural language processing of textual databases has been used to deal with environmental challenges \cite{Bhatia2021,Kim2020,Luccioni2020}, exploration of such databases in Brazilian Portuguese remains underdeveloped, to say the least.

On the dataset availability side, \cite{Paschoal2021} recently released \textit{Pirá}, the first Bilingual Portuguese-English crowdsourced QA dataset about the ocean and, in particular, the Brazilian coast, based on UN reports and abstracts from scientific papers. In English, \textit{CLIMATE-FEVER} is intended to serve as a fact-checking dataset for claims related to climate change \cite{Diggelmann2020}. For topic detection, \textit{ClimaText} is a dataset that leverages manual annotation of labels on texts extracted from Wikipedia and the official U.S. documents via Active Learning \cite{Varini2020}. Furthermore, there is a large gap in the availability of annotated NLP datasets focused on environmental or climate issues, despite the urgency of the topic.

In this work, we start to fill this gap by putting together QA systems that enhance existing architectures and that are built from a knowledge base (KB) consisting of 17K Wikipedia articles in Portuguese\footnote{Articles about the ``Environment of Brazil" taken from the following category of Wikipedia: \url{https://pt.wikipedia.org/wiki/Categoria:Meio_ambiente_do_Brasil}} and 29K news.\footnote{Published between January 2018 to June 2021, and scraped from the three biggest newspapers in circulation in Brazil on different topics related to environmental issues in the country (such as deforestation, the status of indigenous peoples).}
The combination of encyclopedic knowledge and recent news lets us capture the public's changing interests as they are reflected in newspapers.
Due to the scarcity of QA datasets specifically related to environment and climate in Brazil and the high cost of manually generating such a dataset, we have also filtered the PAQ (\textit{Probably Asked Questions}) dataset \cite{Lewis2021}, a massive open-domain QA dataset with 65 million automatically-generated question/answer pairs (QA-pairs); we thus obtained a large set of QA-pairs that was then translated.

We leverage insights from the transformer architecture \cite{Vaswani2017} and self-attention mechanisms \cite{Bahdanau2015}, as well as insights on domain-specific fine-tuning \cite{Raffel2019a}.

Our reader model is based on PTT5 architecture \cite{Carmo2020}, a sequence-to-sequence T5 network~\cite{Raffel2019a} model pre-trained in a corpora in Portuguese. To incorporate the databases in the production of answers, we used the BM25 algorithm~\cite{Robertson2009}, the state-of-the-art in sparse information retrieval, to add context blocks to the question then submitted to T5 for fine-tuning. This dual system, composed of a retrieval module and a neural reader model, has been explored in the QA literature \cite{Guu2020,Lewis2020a} and has some advantages over systems with only a neural module. First, it reduces the document load that the reader needs to process because the retrieval system pre-selects the $k$ best passages in the corpus before submitting them further. Second, the possibility of including new factual information as input from the subsequent reader. Third, it potentially generates less hallucination, a very common problem in text generated by neural networks. 

In short, our main contribution is a QA system focused on the environment in Brazil in Portuguese, developed in two schemes. One scheme consists of only a Reader module containing a language model; the other consists of a Reader equipped with a powerful Retriever with access to documents indicated previously (Wikipedia, news, etc). Another contribution is the filtered and translated corpus of QA-pairs related to our domain; the fundamental idea there was to extract these pairs from a massive open-domain dataset, thus avoiding the manual creation of a new set of QA-pairs\footnote{All relevant code, checkpoints, and data (except for the scrapped news, in respect of copyright laws) are available or referenced in the Github repository: \url{https://github.com/C4AI/deepage}.}. As far as we know, we are the first to design a QA system and a dataset on the Brazilian environment in this way.

In the following sections, we explain how the models work and which settings we used in our experiments. We describe the process of building and filtering the knowledge bases: Wikipedia categories in Portuguese, and newspaper content. We also explain how we enhanced our QA dataset on the environment of Brazil using a massive open-domain QA dataset. Then we present the different experiments we ran, discuss the main results, and analyze possible improvements as future work.

\begin{figure}[t]
\centerline{\includegraphics[width=0.85\textwidth]{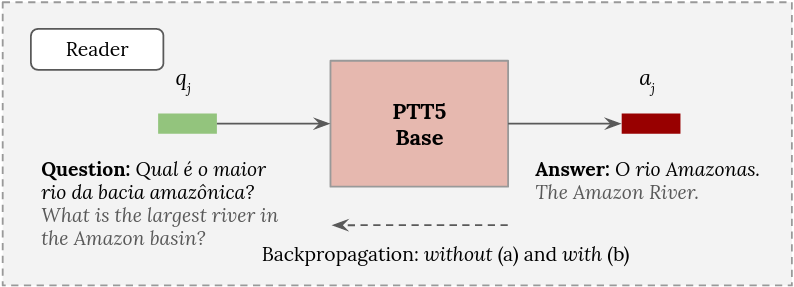}}
\caption{QA system with only PTT5 as a Reader, without a retriever module. Two versions of this reader were tested: \textit{(a)} without fine-tuning, in which the model could only rely on knowledge saved in the network's own weights \textit{(b)} another with fine-tuning on the QA-pair training set.} 
\label{model_t5}
\end{figure}

\begin{figure}[t]
\centerline{\includegraphics[width=0.85\textwidth]{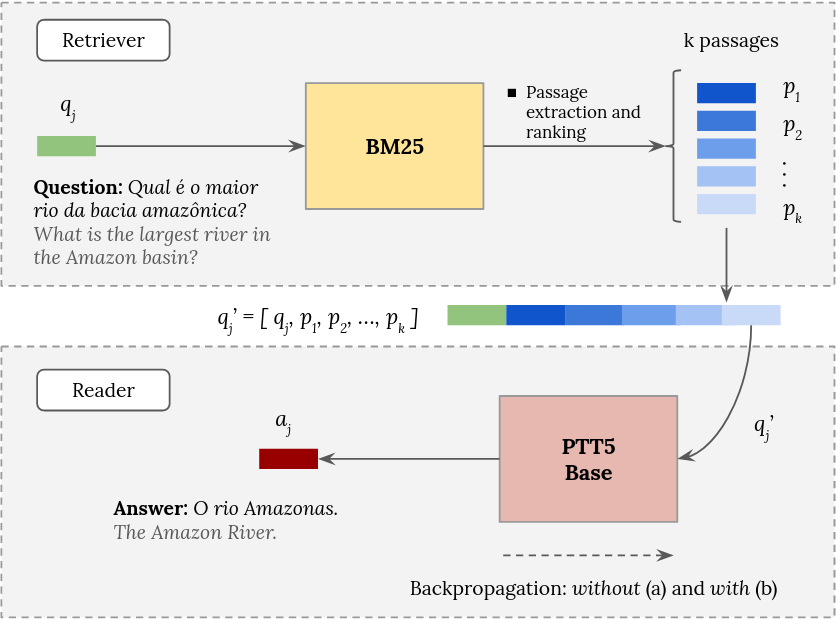}}
\caption{QA system with both the Retriever (BM25) and the Reader (PTT5) modules working together. Again, two versions of this system were tested: \textit{(a)} without fine-tuning \textit{(b)} with fine-tuning on the QA-pair training set.} 
\label{model_bm25_t5}
\end{figure}

\section{Models and Architectures}

In this work, we have two basic model architectures. The first one is illustrated in Figure \ref{model_t5}: without any additional supporting documents, the model answers the posed question by only accessing the information stored in its own parameters. 
We call this architecture a \emph{Reader}-only one. So, for each question $q_j$, the Reader generates another sequence, the answer, $a_j$. The Reader can thus answer questions without text support; in our implementation, it is based on T5, a language model that can generate its own responses from scratch (that is, it does not rely on extracting it from a passage of text).

In the second architecture, we add a \emph{Retriever} module before the Reader. In this case, the Reader has access to an external knowledge base, in addition to the information saved in its parameters. We resorted to Wikipedia and news from mainstream newspapers. For each question $q_j$, the Retriever searches for the most relevant $k$ passages $\{p_1, p_2, ..., p_k\}$ in the corpus. Then the original question is concatenated to these passages, producing a reformulated question $q_j'= [q_j, p_1, p_2, ..., p_k]$, which, finally, serves as input to the Reader, in place of $q_j$, to generate an answer $a_j$. The scheme is represented in Figure \ref{model_bm25_t5}.

In either case, the Reader may or may not be fine-tuned with the QA-pairs $(q_j, a_j)$ from the problem domain. This is expressed by options (b) and (a) presented at the bottom of the Figures \ref{model_t5} and \ref{model_bm25_t5}. 

We next discuss in more detail the technical aspects of the Reader and the Retriever we have implemented.

\subsection{BM25 as the Retriever} 

BM25 \cite{Robertson2009} is an algorithm that estimates the relevance of documents from a set given a query.
It is the state-of-the-art \emph{sparse} retrieval technique, defined as a function of query terms frequency, document length, average document length, and the number of documents containing the query term. 
%%%
We applied BM25 to retrieve sentences of about 100 words from all sentences of the KB. The query is defined as the posed question $q_j$.

\subsection{PTT5 as the Reader} 

PTT5 \cite{Carmo2020}, an encoder-decoder transformer, was pre-trained on the BrWaC Brazilian Portuguese website corpus \cite{WagnerFilho2019} for the task of masked language modeling, where tokens from the corpus are masked so that the model has to predict them. The model is based on T5, which is derived from the original encoder-decoder transformer by Vaswani et al. \cite{Vaswani2017}, characterized by several blocks of self-attention layers concatenated to feed-forward networks.
We applied the ``base" version of PTT5 with a Portuguese vocabulary\footnote{The pre-trained model is available at \url{https://huggingface.co/unicamp-dl/ptt5-base-portuguese-vocab}.}, with 12 layers and 12 attention heads, with a total of 220M trainable parameters.

\begin{figure}[h]
\centerline{\includegraphics[width=0.9\textwidth]{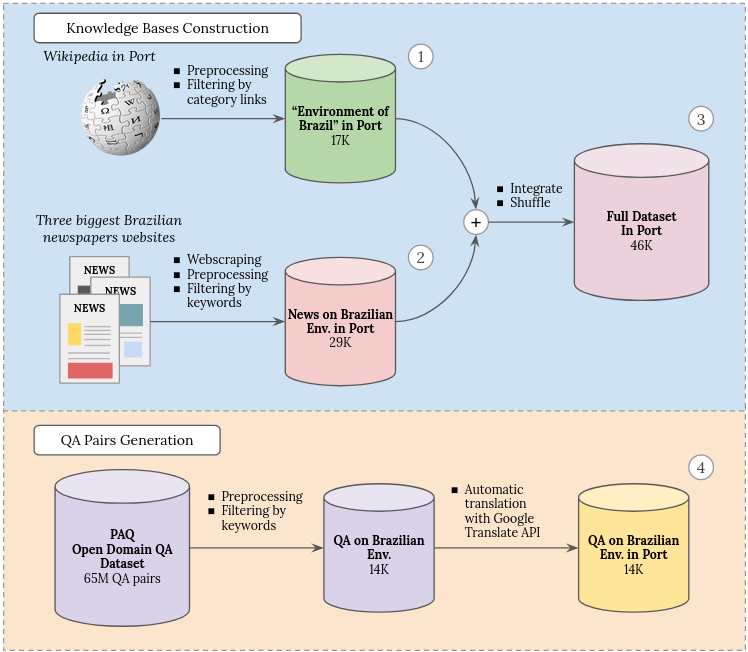}}
\caption{To build the KBs (upper blue section), Wikipedia articles in Portuguese from the category ``Environment of Brazil" were processed, filtered and loaded into a database (1); Newspaper news were filtered by keyword, scraped and also loaded into a database (2). Finally, the two databases were integrated and shuffled to produce a third database (3), the ``full" version. All three were used in different experiments. At the bottom of the figure (orange section), the PAQ filtering process, QA massive open-domain dataset, is described. Using regex with key phrases related to the environment in Brazil, we filtered 14K QA-pairs and then translated them into Portuguese using Google Translation API (4).}
\label{kb_construction}
\end{figure}

\section{Dataset Generation}

We built a dataset based on a set of textual documents and a set of QA-pairs. The KB has two main sources of texts, the Brazilian Portuguese Wikipedia, and newspapers. We describe next how they were collected (as depicted in Figure \ref{kb_construction}).

\subsection{Filtering Articles from Wikipedia in Portuguese}

Wikipedia is divided in such a way that articles are associated with categories. 
A category, on the other hand, is associated with articles and other categories that restrict even more the subject, called subcategories.

To access this information, a SQL table associating article and subcategory identifiers to category names is available for download at Wikimedia's dumps page.\footnote{Available at \url{https://dumps.wikimedia.org/}.}
Thus, to obtain several articles associated with the Brazilian Environment, we applied a recursive script that performs a breadth-first search of articles on subcategories, starting from an arbitrary category title. The algorithm stops when the desired number of articles is reached.

We searched by taking the starting point ``Environment of Brazil" (freely translated from ``Meio Ambiente do Brasil") and obtained 17K Wikipedia articles associated with the subject (hereafter abbreviated as ``Wiki" for simplicity).

\subsection{Scraping and Filtering News from the Biggest Brazilian Newspapers}

To build the news base, we scraped and processed news linked to pre-selected keywords in the three biggest newspapers in the country: \textit{Folha de S.Paulo}, \textit{Estadão} and \textit{O Globo}. Due to the limited value of this type of text over time, we kept only news from January 2018 on, as this is the beginning of the current federal government in Brazil. We downloaded the headline and body of each article and, after final cleaning and pre-processing, we ended up with 29K news (we will refer to this database as ``News").

To select the news that would be scraped, we first carefully crafted a list of keywords that are strongly related to the environment in the country. We then use the native search engines on each newspaper's website to inject these keywords, list the search results, and download them via webscraping techniques. To minimize the number of false positives, i.e. news that is related to a certain keyword but is linked to news from a different subject that is not of our interest, we also excluded articles related to a set of specific words for some keywords. More details on this selection can be found in Appendix B.

\begin{figure}[t]
\begin{center}
\includegraphics[width=0.8\textwidth]{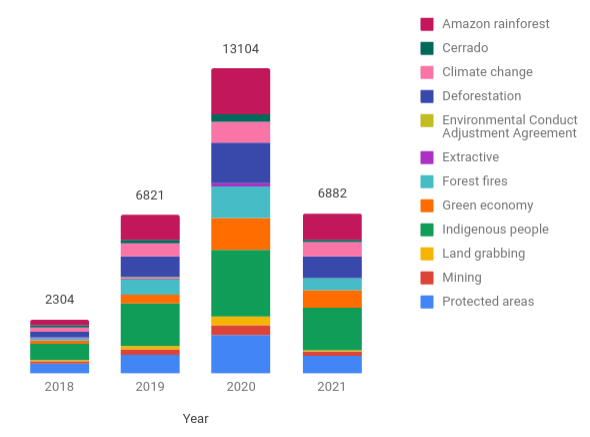}
\caption{Citation count of each environmental category per year since 2018, considering the three biggest newspapers in Brazil.} 
\label{count_keywords_news_pos2018}
\end{center}
\end{figure}

Due to latencies and limitations inherent to the scraping process of websites, there is no guarantee that all news related to a particular term has been downloaded. However, we obtained large numbers related to each keyword, which suggests good coverage, as illustrated in the Figure \ref{count_keywords_news_pos2018}.

\subsection{Filtering and Translating the PAQ QA Dataset}
To obtain an appreciable number of QA-pairs to fine-tune the models, we chose to filter a large-scale open-domain QA dataset with keywords that should and should not be in the questions or answers; this query with multiple rules was carefully hand-crafted and searches were done with regex expressions.\footnote{The query we developed is described in the Appendix A.} 

Initially, we applied this query to filter the MS MARCO v1 \cite{Nguyen2016} training dataset, which is composed of questions from real users made in Bing, with human-annotated answers and contained 80.142 QA-pairs after eliminating unanswered questions. However, due to many constraints we imposed on the filter to avoid false positive and false negative pairs\footnote{For example, if the word ``biome" were a substring of a certain question or its answer, the word ``Brazil" or other names of states in the country should also be so, so as not to include QA-pairs about biomes from other countries in our selection.}, we obtained a return rate of only 0.037\%, which corresponded to 30 pairs, an insufficient amount of data to fine-tune our models. Relaxing the QA-pair filter did not generate a significant increase in this value. 

\begin{figure}[t]
\begin{center}
\includegraphics[width=0.6\textwidth]{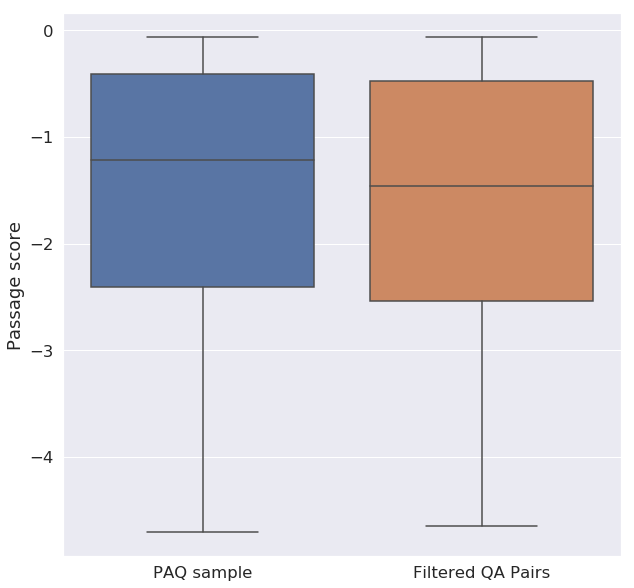}
\caption{Comparison between the distribution of passage scores of a random sample of the PAQ with the ones of QA-pairs filtered for the Brazilian environmental domain. The passage score is a logprob score calculated in the PAQ dataset that measures how likely a given QA-pair is in practice; the closer to zero, the better. Note that the filtering process did not cause a significant loss of quality.}
\label{comp_paqqa_filteredqa}
\end{center}
\end{figure}

Assuming this rate would be similar to other QA datasets based on user queries on search engines, such as Google's Natural Questions \cite{Kwiatkowski2019}, we decided to filter the PAQ dataset, as it would be the only one capable of providing several training pairs of about three orders of magnitude greater than the one obtained from MS MARCO v1, which was what we desired at least. In fact, with a rate of return of about 0.024\%, we got 14,386 QA-pairs after the filter. As shown in the plot \ref{comp_paqqa_filteredqa}, the filtering process did not generate a significant loss of quality for the QA-pairs when compared to the original PAQ.

\begin{figure}[t]
\begin{center}
\includegraphics[width=0.75\textwidth]{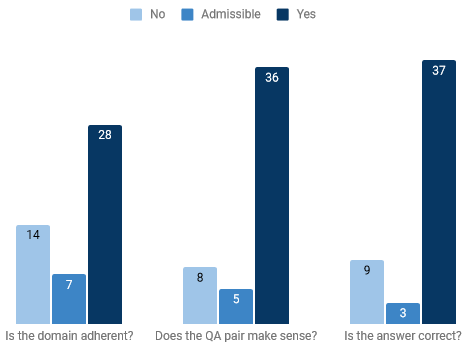}
\caption{Manual evaluation of a sample of 50 QA-pairs from our domain-specific QA dataset, which got at least 70\% of admissible and 50\% perfect results in all three categories} 
\label{manual_evaluation_qas}
\end{center}
\end{figure}

In addition to the quantitative evaluation that demonstrates that our filtered QA-pairs are as good as those in the general PAQ base (which already have high-quality \cite{Lewis2021}), we also performed a manual, qualitative inspection on a sample of 50 instances of our set of QA-pairs. They were evaluated by a human annotator, who answered three questions for each filtered and translated QA-pair: ``Is the domain adherent?", ``Does the QA-pair make sense?" and ``Is the answer correct?". The annotator could answer all questions with only one of the following possibilities: ``Yes", ``Admissible" and ``No". As shown in Figure \ref{manual_evaluation_qas}, the results show that got at least admissible were: 70\% for domain adherence, 82\% for sense, and 80\% for the correctness of the answer, and even considering just perfect QA-pairs, all categories got more than 50\% of occurrence.

Finally, to fine-tune our QA system entirely on the same language, we applied the Google Translate API (Application Programming Interface) to translate these pairs into Portuguese.

\section{Experiments}
To measure the impact of the retrieval module, each base of supporting documents, and the fine-tuning of the system, we performed three groups of experiments:

\begin{enumerate}
\item Reader-only and Retriever+Reader, both without fine-tuning;
\item Reader-only, with fine-tune;
\item Retriever+Reader, with fine-tune.
\end{enumerate}

The filtered QA-pairs dataset was randomly split into 3 groups: 70\% for training, 15\% for validation, and 15\% for the test. The models in experiments 2 and 3, which depended on a training phase, were trained for 30 epochs, with a batch size equal to 16, weight decay equal to 0.01 and a learning rate of 2e-5; the same training and validation sets were used for the fine-tuning. For all models, we report the F1-score, the Exact Match (EM), and the Rouge-L (R-L) metrics, also obtained in the same test set. In all cases, we used the \url{ptt5-base-portuguese-vocab} T5 pre-trained model in Portuguese, since it was the recommended one by the original work \cite{Carmo2020} in comparison to the other versions of PTT5, including its large version. In cases where we used the BM25-based retrieval module, we preprocessed all the supporting documents by removing special characters, eliminating line breaks, and splitting them into chunks of 100-word passages.

\subsection{Experiment 1: Systems Without Fine-Tune}
Experiment 1 aimed to provide a baseline and demonstrate the impact of the lack of a fine-tuning step with the filtered QA base for the problem domain. Therefore, we place the two models, Reader-only and Retriever+Reader, directly to answer the test set questions, without any previous fine-tune, as indicated in Figures \ref{model_t5}(a) and \ref{model_bm25_t5}(a), respectively.

\subsection{Experiment 2: Reader-only, With Fine-Tune}
As in Experiment 1, in this case, we abdicated the retrieval module but performed the fine-tuning of PTT5 in our domain-specific QA dataset. All other parameters are identical to those of the Reader in the previous subsection. Figure \ref{model_t5}(b) illustrates the procedure.

\subsection{Experiment 3: Retriever+Reader With Fine-Tune}
In this experiment, the model was composed of the two modules, Retriever and Reader, as in the second case discussed in Experiment 1, with the difference that now PTT5 is submitted to a fine-tuning on our domain-specific QA dataset, as shown in Figure \ref{model_bm25_t5}(b). In this experiment, we explore the impact on the quality of model answers to:

\begin{enumerate}
\item A larger number of passages retrieved by the retriever;
\item Each KB (Wiki, News and Wiki+News).
\end{enumerate}

Thus, we trained the model with $k = 5$ retrieved passages and 512 entry tokens for PTT5 three times, once with each distinct KB (Wiki, News, and Wiki+News). Then, we repeated the same 3 pieces of training but considering $k = 10$ and 1024 entry tokens for PTT5.

\section{Results and Discussion}
The results of all the experiments described in the previous section are consolidated in Table \ref{mainresults}. As the metrics are highly correlated, we will focus on the F1-score from now on. Information on training and inference times, as well as the machine settings used, can be found in Appendix C.

\begin{table}
\caption{Main results of the tests conducted analysed by the metrics F1-score, Exact Match and Rouge-L. The best model was the one with a reader and a retriever backed only on the Wiki database, with 10 passages retrieved.}
\label{mainresults}
\begin{tabularx}{\textwidth}{|m{2.2cm}|m{4.1cm}|>{\centering\arraybackslash}m{1.9cm}|>{\centering\arraybackslash}m{1.1cm}|>{\centering\arraybackslash}m{1.1cm}|>{\centering\arraybackslash}m{1.1cm}|}
\hline
\textbf{Supporting documents} & \textbf{Model} & \textbf{Number of Passages $k$}  & \textbf{F1} & \textbf{EM} & \textbf{R-L}\\
\hline \hline
None & Reader w/o FT & -  & 3.5  & 0.0 & 3.9 \\
\hline
News + Wiki & Retriever + Reader w/o FT & 5  & 2.9  & 0.0 & 3.2 \\
\hline
None & Reader w/ FT & -  & 32.2  & 24.8 & 32.8 \\
\hline
News & Retriever + Reader w/ FT & 5  & 32.9  & 25.2 & 33.5  \\
\hline
Wiki & Retriever + Reader w/ FT & 5  & 34.7  & 26.9 & 35.4  \\
\hline
Wiki + News & Retriever + Reader w/ FT & 5  & 34.4  & 26.8 & 35.0  \\
\hline
News & Retriever + Reader w/ FT & 10  & 32.2  & 24.9 & 32.7  \\
\hline
Wiki & Retriever + Reader w/ FT & 10  &  \textbf{36.2}  & \textbf{28.2} & \textbf{37.0} \\
\hline
Wiki + News & Retriever + Reader w/ FT & 10  & 35.4  & 27.5 & 36.2  \\
\hline
\end{tabularx}
\end{table}

\subsection{Importance of Fine-Tune on the Specific Domain}
As expected, the models from Experiment 1, without fine-tune, had the worst results of all. This demonstrates the importance of this training phase and the construction of the domain-specific dataset we performed: the Reader-only from Experiment 2, fine-tuned in our QA dataset, performed 11 times better than the Reader+Retriever system from Experiment 1, which had access to a KB composed of all the news and Wikipedia articles collected, and achieved scores comparable to models that were fine-tuned and had access to one KB. Then, in short, the presence of the retrieval module does not compensate for the absence of the model's fine-tune.

\subsection{Impact of Different KBs on Scores}
Experiment 3 allowed us to compare the effect of each KB on the quality of the systems' responses with the Retriever. For both $k = 5$ and for $k = 10$, we observed that the systems supported \textit{only} by the Wiki KB performed better than those that have access to the expanded KB with newspaper news (Wiki+News), which was contrary to what we expected at the beginning. Also, in both cases, the least competitive results occurred when the KB is composed only by the News KB, still, however, surpassing the Reader-only model. 

A possible explanation for this phenomenon may lie in the PAQ construction process, which is automatically generated on Wikipedia passages in English. This can generate a bias in favor of the KB formed by a specific category of Wikipedia, even though it is, here, in Portuguese.

\subsection{Influence of Distractors on the Reader}
Finally, it is remarkable that doubling the number of retrieved passages to $k = 10$ improves the performance of the models with Wiki, even when integrated with the News, but simultaneously the worst model among those that have access to a KB is the one composed \textit{solely of news}. It is slightly inferior even to the same model configured with $k = 5$. Arguably, this is because, when we concatenate the initial question $q$ with the 10 passages instead of 5, we end up diluting the weight of $q$ in the reformulated question $q'$ (a much longer text), making the original question more diffuse among heterogeneous news articles, which does not compensate for the information gain brought by the retrieved passages $\{p_1, p_2, ..., p_k\}$. Still, the News KB can be helpful due to the contemporaneity it aggregates to the KB, which can be particularly relevant for reasons of explainability.

The same probably does not occur as often with the Wiki KB perhaps because of the generational bias of PAQ. Nevertheless, it does not make it immune to distractors. Table \ref{docs_impact} illustrates two emblematic cases, comparing a QA system with only the Reader and another with the Retriever+Reader. In the first question, whose correct answer is ``2001", the Reader-only was wrong, but the system composed also of a Retriever was right. The retrieved passage that contains the answer is: 

\begin{quote} 
``\textit{...Today Fernando de Noronha's economy depends on tourism (...) \textbf{In 2001 the archipelago was declared a World Heritage Site}, including the Atol das Rocas, as Sítio das Ilhas..."}. 
\end{quote}

Thus, we see that the information extracted by the Retriever was essential for the Reader not to incur the same error again.

However, in the second question, the opposite occurs: the Reader gets it right, but the Retrieve+Reader system gets it wrong. When observing the recovered passages, we noticed a potential distractor that could have induced the system to error: 

\begin{quote}
\textit{...On March 8 of that year, Marabá was practically submerged. Occupying an area of 803 250 square kilometers, it is the largest hydrographic basin entirely in Brazil, even though it belongs to the Amazon Basin (...) \textbf{The Tocantins, the main river in this basin}, rises in the north of Goiás and flows into the Marajoara Gulf...}". 
\end{quote}

Despite these occasional problems, all tests we performed indicated that a system consisting of a Retriever and a Reader always surpasses one with only a Reader.

\begin{table}[t]
\caption{Comparison between a case in which the presence of the Retriever is useful to prevent the Reader from making mistakes (first case), and another in which the retrieved passages otherwise mislead it due to the presence of distractors.}
\label{docs_impact}
\begin{tabularx}{\textwidth}{|m{4.3cm}|m{2.2cm}|m{2.2cm}|m{3.03cm}|}
\hline
\textbf{Question} & \textbf{True Answer} & \textbf{Reader-only} & \textbf{Retriever+Reader }\\
\hline \hline
\makecell[cl]{Quando Fernando de Noronha \\ se tornou um Patrimônio \\Mundial da UNESCO?\\ \textit{(When did Fernando de}\\ \textit{Noronha become a UNESCO} \\\textit{World Heritage Site?})} & 2001 & 1997 & \textbf{2001} \\
\hline
\makecell[cl]{Qual é o maior rio da bacia \\amazônica? \\\textit{(What is the largest river in} \\\textit{the Amazon basin?)}} & \makecell[cl]{Rio \\Amazonas \\(\textit{Amazon river})} & \makecell[cl]{\textbf{Rio} \\\textbf{Amazonas} \\\textit{(Amazon river)}} & \makecell[cl]{Tocantins\\\textit{(Tocantins)}} \\
\hline
\end{tabularx}
\end{table}

\section{Conclusion}
We presented the first QA system focused on environmental issues in Brazil ---
more importantly, a QA system based on the Portuguese language, a language
that has received remarkably low attention when it comes to automatic question
answering. We combined PTT5 as the Reader, the state-of-the-art among pre-trained language models, and BM25 as the Retriever, the state-of-the-art sparse retrieval technique. Also, we collected documents and QA-pairs by filtering articles related to ``Environment of Brazil" in the Wikipedia dump in Portuguese; scraped environmental news from January 2018 to June of 2021 in the three most important newspapers in the country; filtered and translated a recently released massive open-domain QA dataset to obtain a substantial domain-specific set of QA pairs. Despite potential generation biases found in this last step, our trained QA systems demonstrated competitive scores. We hope that this work can stimulate similar initiatives on a topic that is so relevant to Brazilian environmental efforts.

To DEEPAGÉ increase social awareness and understanding about the environment and the climate of Brazil, the system must be tested with human subjects. Integration with other modules such as a social chatbot can certainly make the system more appealing for users \cite{Gao2018}. Another further improvement would be the construction of a system that gives more complete and elaborate answers. As our training was ran using the PAQ dataset, with a majority of factual and short responses, the system is not prepared to give long and detailed answers. Also, generative models such as the PTT5 can hallucinate when giving answers, especially when there is no retriever. Hence, another useful future extension to DEEPAGÉ would be a filtering module to avoid absurd answers.

\section*{Acknowledgements}
This work was financed in part by the \textit{Coordenação de Aperfeiçoamento de Pessoal de N\'{i}vel Superior} (CAPES, Finance Code 001) and by the \textit{Ita\'{u} Unibanco S.A.}, through the \textit{Programa de Bolsas Ita\'{u}} (PBI) of the \textit{Centro de Ciência de Dados} (C$^2$D) of \textit{Escola Polit\'{e}cnica} of \textit{Universidade de São Paulo} (USP). We also gratefully acknowledge support from \textit{Conselho Nacional de Desenvolvimento Cienti\'{i}fico e Tecnol\'{o}gico} (CNPq) (grants 312180/2018-7 and 310085/2020-9) and the \textit{Center for Artificial Intelligence} (C4AI-USP), with support by the \textit{Fundação de Amparo à Pesquisa do Estado de São Paulo} (FAPESP, grant 2019/ 07665-4) and by the \textit{IBM Corporation}.
\bibliographystyle{splncs04}
\bibliography{blab_answerer-bracis21}

\section{Appendix A}
To filter QA-pairs in PAQ, we created four sets of keywords: $M$ (from ``Must have": ``brazil*" and other states names), $G$ (from ``Good to have"; e.g. ``deforestation"), $U$ (from ``Unique expressions"; e.g. ``ibama" or ``amazon rainforest") and $E$ (from ``to Exclude"; e.g. ``soccer"). For a pair to be selected, it must contain, in its question or answer, at least one expression belonging to $M$ or $U$. If it was from $U$, it would already be selected; if it was from $M$, it could not contain any $E$ expressions. Finally, if the QA-pair contains any keyword from $G$, it should also contain at least one from $M$, but none from $E$ either.

\section{Appendix B}
We performed queries associated with the following keywords (here, translated into English): \textit{Amazon rainforest, Cerrado, Climate change, Deforestation, Environmental Conduct Adjustment Agreement, Extractive} (from ``\textit{Extractivism}"), \textit{Forest fires, Funding for conservation, Green economy, Land grabbing, Mining} and \textit{Protected areas}. It should be noted that the decision to exclude some words is only understandable in Portuguese, since there is no reasonable parallel in English, as with the word ``fist" for the keyword ``Cerrado" - in Portuguese, the expression ``punho cerrado" (``clenched fist") is common, but it is clearly not directly related to the Cerrado biome. Also for more significant results, we eliminated news related to ``agrobusiness" in the \textit{Green economy} keyword and ``militias" in the \textit{Land grabbing} one.

\section{Appendix C}
To perform the experiments, we leverage a machine with an AMD Ryzen 9 3950X Processor with 32 CPUs, 64 GB of RAM, 2 NVIDIA RTX 3090 GPUs of 24GB each, on an Ubuntu 20 LTS. Under these conditions, the training of each model lasted between 5h and 8h, with the best model being trained in about 7h40. The inference time -- that is, the time it took to a model to generate an answer to a single question from the test set -- was around 0.006s.

\end{document}